\documentclass[conference]{IEEEtran}
\IEEEoverridecommandlockouts
\usepackage{cite}
\usepackage{amsmath,amssymb,amsfonts}
\usepackage{algorithmic}
\usepackage{graphicx}
\usepackage{textcomp}
\usepackage{xcolor}
\usepackage{booktabs}
\usepackage{hyperref}
\hypersetup{
    colorlinks=true,
    linkcolor=blue,
    filecolor=magenta,      
    urlcolor=cyan,
    pdftitle={Overleaf Example},
    pdfpagemode=FullScreen,
    }

\def\BibTeX{{\rm B\kern-.05em{\sc i\kern-.025em b}\kern-.08em
    T\kern-.1667em\lower.7ex\hbox{E}\kern-.125emX}}
\begin{document}

\title{MMIS: Multimodal Dataset for Interior Scene Visual Generation and Recognition}

\author{\IEEEauthorblockN{ Hozaifa Kassab, Ahmed Mahmoud, Mohamed Bahaa, Ammar Mohamed, Ali Hamdi}
\IEEEauthorblockA{\textit{MSA University} \\
 Giza, Egypt \\
\{hozaifa.fadl,ahmed.mahmoud60, mohamed.bahaa4,ammohammed,ahamdi\}@msa.edu.eg }}

 
\IEEEoverridecommandlockouts
\IEEEpubid{\makebox[\columnwidth]{979-8-3503-6263-6/24/\$31.00 ©2024 IEEE \hfill}
\hspace{\columnsep}\makebox[\columnwidth]{ }}
\maketitle
\IEEEpubidadjcol

\begin{abstract}
We introduce MMIS, a novel dataset designed to advance \textit{M}ulti-\textit{M}odal \textit{I}nterior \textit{S}cene generation and recognition. \textit{MMIS} consists of nearly $160,000$ images. Each image within the dataset is accompanied by its corresponding textual description and an audio recording of that description, providing rich and diverse sources of information for scene generation and recognition. \textit{MMIS} encompasses a wide range of interior spaces, capturing various styles, layouts, and furnishings. To construct this dataset, we employed careful processes involving the collection of images, the generation of textual descriptions, and corresponding speech annotations. The presented dataset contributes to research in multi-modal representation learning tasks such as image generation, retrieval, captioning, and classification. The dataset is available at the following URL: \url{https://github.com/AhmedMahmoudMostafa/MMIS}.
\end{abstract}

\section{Introduction}
Multi-modal deep learning\cite{akkus2023multimodal} is an active, multi-disciplinary research field that encompasses a diverse range of disciplines focused on creating intelligent computer systems capable of understanding, reasoning, and learning from various types of information from multiple sources or modalities, such as images, text, audio, and sensor data\cite{liang2023foundations,hamdi2021marl}. The field has gained traction in recent years, particularly with the growing interest in multimodal tasks such as text-to-image generation\cite{bie2023renaissance}, text-image retrieval\cite{cao2022imagetext}, image captioning\cite{vinyals2015tell}, and visual question answering\cite{agrawal2016vqa}. By combining insights from various modalities, multi-modal deep learning models can capture rich and nuanced patterns that may not be discernible from any single modality alone\cite{bioengineering11030219}. One of the driving forces behind advancements in multi-modal deep learning is the availability of large-scale datasets that encompass multiple modalities, such as images, text, and speech. These datasets not only facilitate the development and evaluation of novel and robust models but also pave the way for interdisciplinary research aimed at leveraging the synergies between different data types.

\begin{figure}{}
    \centering
    \includegraphics[width=0.5\textwidth]{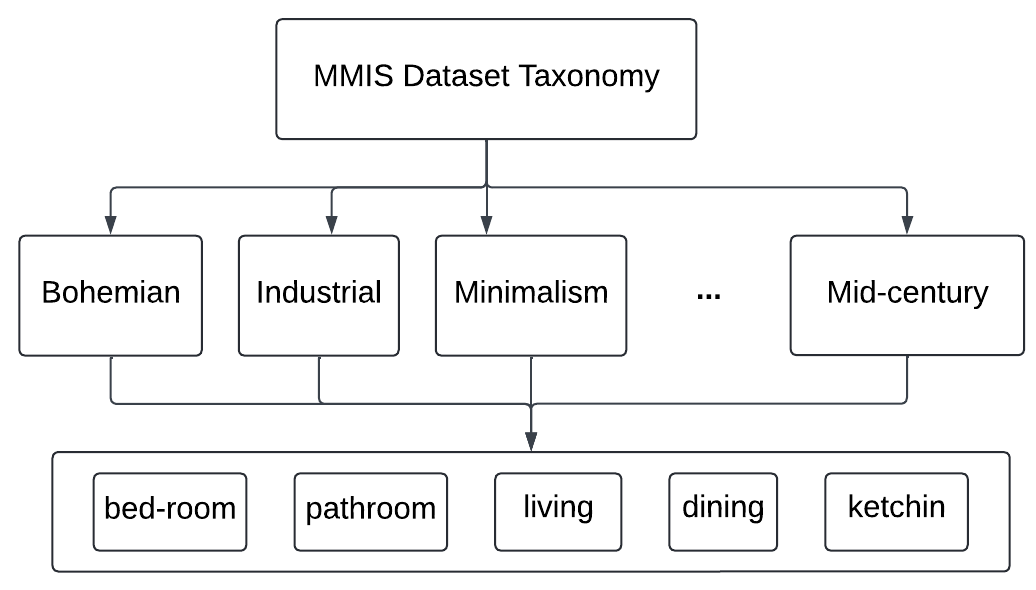}
    \caption{Dataset Taxonomy}\vspace{-5mm}
\end{figure}

In this paper, we introduce a novel dataset specifically designed to explore the fusion of multiple modalities for various tasks: image generation, retrieval, and captioning. The dataset, composed of images, textual descriptions, and corresponding speech samples, offers a rich resource for investigating the complementary nature of different modalities in capturing and representing complex visual scenes. Our dataset focuses on interior design images, a domain that presents unique challenges and opportunities for multi-modal analysis due to its rich visual and semantic content. The motivation behind creating this dataset stems from the growing interest in multi-modal AI systems that can understand and generate content across different modalities. While significant progress has been made in individual modalities such as image processing, text analysis, and speech recognition, there remains a gap in our understanding of how these modalities can be effectively integrated to solve real-world problems. By providing a diverse collection of images along with textual descriptions and speech annotations, we aim to facilitate research into multi-modal learning techniques that can leverage the complementary strengths of each modality. Our dataset comprises images representing various styles of interior design, ranging from modern and minimalist to bohemian and traditional. Each style category includes images depicting five different attributes: bedroom, living room, bathroom, kitchen, and dining room. The dataset is structured such that each attribute contains a variable number of images, with counts ranging from $600$ to $1,200$ images per attribute. Additionally, each image is accompanied by a textual description and an audio recording of the description, providing multiple modalities of data for analysis. While our dataset can be utilized for a wide range of tasks, including image classification, captioning, and speech recognition, our primary focus lies in exploring its potential for image generation and retrieval tasks. By leveraging the unique characteristics of our dataset, we aim to explore how different techniques for fusing information from images, text, and speech can enhance the performance of image generation and retrieval tasks. Additionally, we seek to investigate methods for learning joint representations across various modalities to enable seamless information exchange and integration. These endeavors will aid in developing models capable of comprehending the semantic content of images through the simultaneous analysis of visual, textual, and auditory cues. To ensure comprehensive assessment, we have defined appropriate evaluation metrics and benchmarks for assessing the performance of multi-modal systems on our dataset.

\begin{figure*}[h]
    \centering
    \includegraphics[width=1\textwidth]{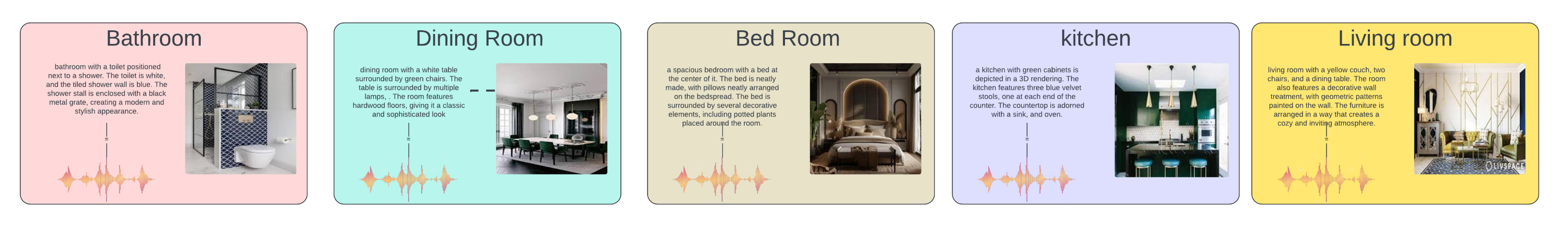}
    \caption{Sample from the interior design styles the Art Deco Style with the five Rooms }
    \label{fig:example}
\end{figure*}

\section{Related Work}
In this section, we review existing datasets relevant to our multi-modal dataset, focusing on three main categories: image datasets, image-text datasets, and image-text-audio datasets. These datasets have played a crucial role in advancing research in machine learning tasks, including image classification, captioning, and generation. Table II summarizes some representative datasets and their specific statistics.
\subsection{ \textbf{Image Datasets}}Image datasets, such as ImageNet, CIFAR-10 \cite{cifar10}, and Oxford-102 Flowers, serve as foundational resources for a wide range of computer vision tasks. These datasets provide labeled images across diverse categories, enabling researchers to train and evaluate machine learning models for tasks such as image classification, object detection, and scene recognition. For instance, in image classification, where the goal is to assign a single label to an image from a predefined set of categories, these datasets offer a large and varied collection of images with corresponding ground truth labels, allowing researchers to develop and benchmark classification algorithms. Similarly, in object detection tasks, where the goal is to identify and localize multiple objects within an image, they provide annotations specifying the bounding boxes or segmentation masks for each object instance, facilitating the training and evaluation of object detection models. Moreover, scene recognition tasks, which involve categorizing images based on their overall scene category (e.g., indoor vs. outdoor), benefit from the diverse scene categories and annotations provided by these datasets. We will explore our proposed MMIS dataset, which contains images from different classes and labels for those images.
\\
TThe \textbf{ImageNet \cite{5206848}} dataset comprises $14,197,122$ annotated images organized according to the WordNet hierarchy. Since $2010$, it has been utilized in the ImageNet \cite{5206848} Large Scale Visual Recognition Challenge (ILSVRC), serving as a benchmark for tasks like image classification and object detection. This dataset includes manually annotated training images and a separate set of test images with withheld annotations. The \textbf{CIFAR-100 \cite{cifar10}} dataset, a subset of the Tiny Images dataset, contains \$60,000\$ color images sized $32 \times 32$ pixels. It encompasses 100 classes, which are further grouped into 20 super-classes. Each class comprises $600$ images, with both "fine" labels and "coarse" labels. The dataset includes 500 training images and 100 testing images per class. It is the same as \textbf{CIFAR-10 \cite{cifar10}} but it is categorized into 10 classes. The \textbf{MNIST} database is a large collection of handwritten digits. It has a training set of $60,000$ examples, and a test set of $10,000$ examples. It is a subset of a larger NIST Special Database 3 and Special Database 1, which contain monochrome images of handwritten digits. The digits have been size-normalized and centered in a fixed-size image. The \textbf{Oxford 102 Flower} dataset is designed for image classification and comprises 102 categories of flowers commonly found in the United Kingdom. Each category contains between 40 and 258 images. The dataset is challenging due to variations in scale, pose, and lighting across images. However, the proposed MMIS establishes a new challenge to classify those sets of related scenes and styles, which makes it a powerful measure for classification algorithms.

\subsection{\textbf{Image-Text Datasets}}
Image-text datasets, such as MSCOCO \cite{lin2015microsoft}, CelebA-HQ, Visual Genome, and Fashion-Gen Dataset, offer rich resources for exploring the interaction between visual and textual modalities. These datasets provide paired images and textual descriptions, enabling researchers to tackle tasks that require understanding and generating content across both modalities. One key application is image generation, where the goal is to synthesize images from textual descriptions. By leveraging the paired image-text data, researchers can achieve this by utilizing techniques such as generative adversarial networks (GANs) and transformer models to translate text inputs into coherent visual outputs. Additionally, these datasets are valuable for tasks such as visual question answering (VQA), where models are tasked with answering questions about the content of images using both visual and textual cues. Furthermore, researchers can utilize these datasets for tasks like image retrieval, where the textual descriptions serve as additional semantic cues to improve the accuracy and relevance of retrieved images.
 The \textbf{Microsoft COCO \cite{lin2015microsoft}} dataset comprises over $200,000$ high-quality images, each accompanied by multiple human-annotated captions, providing detailed descriptions of the visual content. Annotations include bounding boxes and segmentation masks, ensuring spatial understanding of objects. The dataset's diversity spans various scenes, objects, and activities, facilitating robust model training \cite{benbaruch2021asymmetric}. The \textbf{VQA} dataset, referred to as VQA dataset19, is tailored for open-ended questions related to images. It comprises $265,016$ images, with each image associated with a minimum of 3 questions (averaging 5.4 questions) and 10 answers provided for each question. The \textbf{Fashion-Gen} dataset is a recently introduced collection consisting of $293,008$ high-definition fashion images, each sized at $1360 \times 1360$ pixels. Additionally, this dataset includes item descriptions meticulously crafted by professional stylists, enhancing its utility for various fashion-related tasks. The \textbf{Image-Chat} dataset is a component of a vast conversational database, containing hundreds of millions of examples. Specifically, it comprises $202,000$ dialogues and $401,000$ utterances, all associated with $202,000$ images. Additionally, this dataset incorporates 215 possible personality traits, enriching the conversational context for analysis and modeling purposes. Our proposed MMIS contains a detailed description of the different interior design styles, focusing on the semantic meaning of the images and the details of the spatial arrangement of objects, which makes it suitable for image generation and retrieval tasks.

\subsection{\textbf{Image-Text-Audio Datasets}}
While initially focused on images and text, recent developments have seen the inclusion of audio modalities in datasets like \textbf{CUB-200} \cite{He_2020}, \textbf{MIT Sub-places}, and \textbf{Flicker30k}. Although the original versions of these datasets were devoid of audio annotations, researchers have since expanded them to incorporate audio descriptions for specific subsets. These additions have significantly broadened the scope of multimodal research, especially in tasks like image generation and retrieval. By combining images, text, and now audio, these datasets provide a rich resource where each modality offers complementary insights, enhancing the overall understanding and representation of the underlying concepts. For instance, in CUB-200, the introduction of textual descriptions alongside images of bird species facilitates a deeper semantic comprehension of the depicted birds, despite the absence of original audio annotations. Similarly, MIT Sub-places now includes audio descriptions paired with images, thereby advancing research in audio-visual scene understanding, a feature not present in the dataset's initial release. Flicker30k, originally focused on images and textual descriptions, has expanded to accommodate audio annotations, allowing researchers to explore the alignment between visual, textual, and auditory cues.
 
\begin{table}
    \centering
    \begin{tabular}{p{1.5cm}ccccc}
        \toprule
        Style & Bedroom & Bathroom & Dining Room & Living & Kitchen \\
        \midrule
        Art-Deco & 627 & 321 & 745 & 713 & 788 \\
        Bohemian & 597 & 634 & 687 & 583 & 703 \\
        Coastal & 719 & 824 & 792 & 504 & 894 \\
        Transitional & 816 & 845 & 806 & 842 & 782 \\
        Mediterranean & 748 & 631 & 630 & 756 & 865 \\
        Eclectic & 861 & 858 & 828 & 851 & 850 \\
        Rustic & 714 & 864 & 821 & 803 & 947 \\
        Mid-Century & 702 & 790 & 678 & 810 & 800 \\
        Farmhouse & 642 & 802 & 794 & 805 & 803 \\
        Japanese Zen & 721 & 937 & 866 & 736 & 712 \\
        Mediterranean & 1005 & 783 & 607 & 820 & 786 \\
        Victorian & 1046 & 984 & 921 & 929 & 600 \\
        French & 1002 & 852 & 882 & 751 & 870 \\
        Tropical & 853 & 980 & 714 & 978 & 783 \\
        Southwestern & 882 & 900 & 1000 & 980 & 780 \\
        Gothic & 642 & 988 & 683 & 720 & 975 \\
        Nouveau & 1028 & 783 & 691 & 714 & 1011 \\
        Retro & 707 & 704 & 1025 & 744 & 849 \\
        Bauhaus & 668 & 821 & 904 & 784 & 1028 \\
        Nautical & 936 & 721 & 711 & 637 & 769 \\
        Hollywood & 850 & 1048 & 989 & 814 & 669 \\
        Moroccan & 626 & 684 & 753 & 877 & 1093 \\
        Tribal & 920 & 889 & 1025 & 607 & 996 \\
        Asian-inspired & 904 & 867 & 951 & 1057 & 790 \\
        Urban & 863 & 825 & 740 & 826 & 1055 \\
        Cottagecore & 786 & 695 & 605 & 786 & 1040 \\
        Contemporary-classic & 990 & 669 & 1036 & 872 & 670 \\
        Retro & 763 & 682 & 723 & 772 & 1050 \\
        Zen & 764 & 693 & 658 & 962 & 1082 \\
        Steampunk & 827 & 853 & 1080 & 818 & 704 \\
        Space-saving & 988 & 967 & 994 & 870 & 1023 \\
        High-tech & 644 & 642 & 1053 & 884 & 925 \\
        Neoclassical & 1062 & 1049 & 1046 & 637 & 1043 \\
        Scandinavian & 831 & 765 & 1074 & 608 & 1057 \\
        Neo-tradition & 1080 & 941 & 848 & 671 & 804 \\
        \bottomrule
    \end{tabular}
    \caption{Dataset statstics}\vspace{-5mm}
    \label{tab:room_widths}
\end{table}

\section{\textbf{Methodology}}
In this section, we outline the steps taken to create and refine the dataset. We explain how we collected, cleaned, pre-processed, and annotated the data, ensuring its quality and reliability. Our goal is to provide a clear understanding of the process involved in integrating images, textual descriptions, and audio recordings into the dataset. By detailing each phase of the methodology, we aim to create a dataset that can be used effectively for research in multi-modal scene understanding and related areas.

\subsection{\textbf{image data}}
The initial phase of dataset creation involved a systematic approach to sourcing images from various online platforms. Recognizing that the detailed layout of any apartment typically encompasses five primary rooms—living room, bedroom, dining room, bathroom, and kitchen—we structured our search methodology around these fundamental spaces. Additionally, we acknowledged the diverse spectrum of interior design styles prevalent in contemporary spaces, ranging from minimalist and Moroccan to bohemian and modern. To ensure comprehensive coverage, we conducted focused searches for each style in conjunction with the five rooms, thereby facilitating a holistic representation of interior design aesthetics. Subsequently, leveraging automated web scraping techniques, we systematically extracted images from reliable online sources, ensuring adherence to copyright regulations and ethical considerations. The MMIS dataset was then subjected to cleaning procedures to eliminate duplicates, low-quality representations, and irrelevant content, thereby refining its composition to include only high-fidelity, pertinent imagery. Additionally, recognizing the importance of standardization for downstream processing tasks, all collected images were resized to a uniform dimension of $256 \times 256$ pixels, ensuring consistency across the dataset. By organizing and standardizing the dataset through a combination of systematic search, automated extraction, and rigorous cleaning procedures, we ensured the integrity and reliability of the collected dataset. This curated dataset forms the foundation for subsequent annotation and pre-processing stages, enabling comprehensive analysis and exploration of multi-modal scene understanding tasks.\vspace{-1mm}

\subsection{\textbf{ Text Data}}
Image captions play a pivotal role in enriching the semantic understanding of visual data, providing textual descriptions that complement and contextualize the associated images. In the context of our dataset, captions serve as invaluable annotations, offering insights into the spatial arrangement, stylistic elements, and functional attributes of interior design compositions. By associating each image with a descriptive caption, we augment the dataset with rich, multi-modal information, enabling deeper comprehension and analysis of the underlying visual scenes. To automate the process of caption generation, we employed the LLaVA v2 model—a state-of-the-art architecture tailored for visual question-answering (VQA) tasks. The LLaVA v2 model leverages advancements in deep learning and natural language processing to generate coherent and contextually relevant captions in response to queries posed about the interior design depicted in the images. This approach not only streamlines the annotation process but also enhances the interpretability and accessibility of the dataset, facilitating diverse applications ranging from image retrieval to content generation. At its core, the LLaVA v2 model harnesses a transformer-based architecture, known for its ability to capture long-range dependencies and semantic relationships in sequential data. Through a process of self-attention mechanisms, the model dynamically weighs the importance of different regions within the input image and integrates this information with contextual cues from the query text. By iteratively attending to relevant image features and refining its predictions based on feedback signals, the model generates captions that are tailored to the specific attributes and nuances of the interior design depicted in each image. By integrating the LLaVA v2 model into the dataset creation pipeline, we not only streamline the annotation process but also enhance the richness and granularity of the dataset annotations. The resulting dataset, enriched with descriptive captions generated by the LLaVA v2 model, serves as a valuable resource for advancing research in multi-modal interior scene generation and recognition, image captioning, and related tasks.

\subsection{ \textbf{Audio Data} }
Incorporating audio data into the MMIS allows for a more comprehensive understanding and interpretation of interior design scenes. To generate speech corresponding to the textual descriptions associated with each image, we utilized the Multi-Speaker Neural Text-to-Speech model—a cutting-edge architecture designed to synthesize high-quality speech with various voices from a single model. The importance of audio data lies in its ability to provide auditory context and sensory cues that complement the visual and textual aspects of the dataset. By incorporating speech synthesis capabilities, we enhance the accessibility and inclusivity of the dataset. Additionally, audio annotations offer valuable supplementary information, enabling a deeper analysis and understanding of different data composition techniques. The Multi-Speaker Neural Text-to-Speech model introduces a novel technique for augmenting neural text-to-speech synthesis with low-dimensional trainable speaker embeddings, allowing for the generation of different voices from a single model. This approach builds upon the advancements made in single-speaker neural TTS models such as Deep Voice 1 and Tacotron, demonstrating improvements in audio quality and speaker diversity. The Multi-Speaker Neural Text-to-Speech model demonstrates the capability for multi-speaker speech synthesis, allowing a single neural TTS system to learn hundreds of unique voices from minimal data per speaker while preserving speaker identities almost perfectly. The model synthesizes high-quality speech with diverse voices, enriching the dataset with a spectrum of auditory experiences. By integrating audio data into the dataset, we enhance the richness and diversity of the multi-modal annotations, facilitating advanced research and applications in multi-modal scene understanding and related domains.

\begin{table*}[htbp]
    \centering
    \begin{tabular}{@{}lcccccc@{}}
        \toprule
        \textbf{DataSet} & \textbf{No. of images} & \textbf{Image resolution} & \textbf{No. of classes} & \textbf{Image} & \textbf{Text} & \textbf{Audio} \\
        \midrule
        COCO & 123,287 & Arbitrary & \_ & Yes & yes & no \\
        Flickr & 100,000 & Variable & - & Yes & yes & no \\
        MIT-places & 10,000 & Variable & - & Yes & yes & No \\
        COCO & 330,000 & Variable & 80 & Yes & No & No \\
        CIFAR & 60,000 & 32x32 & 10 & Yes & No & No \\
        \textbf{MMIS (Ours)} & 200,000 & 256x256 & 30 & Yes & Yes & Yes \\
        \bottomrule
    \end{tabular}
    \caption{Comparison in terms of modalities availability with common Datasets}\vspace{-5mm}
    \label{tab:dataset_info}
\end{table*}

\section{\textbf{MMIS Dataset}}
Our multi-modal dataset, termed "MMIS," is specifically curated to facilitate research and development in multimodal deep learning. The dataset comprises a total of $40$ different classes, each representing a distinct interior design style. These styles encompass a wide spectrum of aesthetics, ranging from modern and minimalist to classic and eclectic. Each class within the dataset is further subdivided into five distinct rooms commonly found in interior spaces: living room, bedroom, dining room, bathroom, and kitchen. The dataset is meticulously curated to ensure diversity and comprehensiveness, with each class containing a variable number of images ranging from $500$ to $1,000$. In total, the dataset comprises approximately $200,000$ samples, providing a rich and extensive resource for training, validation, and evaluation purposes. For each image in the dataset, accompanying textual descriptions are provided to capture the semantic content and contextual details of the interior space depicted. Additionally, speech annotations corresponding to the textual descriptions are included.

\textbf{Table-1} This section summarizes the key statistics of the MMIS dataset, including the number of classes, rooms per class, images per room, and the total number of samples. These statistics provide insights into the composition and scale of the dataset, highlighting its richness and diversity as a resource for interior design research and development.

\section{\textbf{Benchmark}}
To thoroughly assess the quality and versatility of our interior design multimodal dataset, we designed and conducted an extensive suite of benchmark experiments, focusing on two primary machine learning tasks: classification and image generation.

\begin{table}[h]
\centering
\begin{tabular}{@{}lcc@{}}
\toprule
\textbf{Models} & \textbf{Without fine-tuning} & \textbf{With fine-tuning} \\
\midrule
VGG16           & 42\% & 82\% \\
VGG19           & 44.7\% & 87\% \\
RESNET18        & 42\% & 80\% \\
RESNET34        & 40\% & 82\% \\
RESNET50        & 42\% & 84\% \\
EfficientNetB0  & 45\% & 88\% \\
densenet121     & 42\% & 83\% \\
densenet169     & 44.6\% & 88\% \\
densenet201     & 45\% & 85\% \\
\bottomrule
\end{tabular}
\caption{Classification benchmark on the five categories"Rooms"}\vspace{-2mm}
\label{tab:classification}
\end{table}

\subsection{Classification Task}
The classification task is crucial in machine learning, requiring models to accurately categorize data into predefined classes. For this benchmark, we leveraged our dataset's diverse range of interior design styles and categories to evaluate a state-of-the-art classification model's ability. We curated a balanced subset of the dataset across 35 styles and their respective room categories, with nearly 65k images. The classification \cite{srivastava2023omnivec} is done for the five main categories: Bedroom, Dining Room, Living Room, Bathroom, and Kitchen. Eight different pretrained models were selected as the baseline architectures due to their proven performance in visual recognition tasks. These models are VGG16, VGG19, ResNet-18, ResNet-34, ResNet-50, DenseNet121, DenseNet169, and DenseNet201. Our training pipeline included standard data augmentation strategies, such as random cropping and horizontal flipping, to improve generalization. Cross-entropy loss and an adaptive learning rate were utilized for training, and performance was measured using the top-1 accuracy metric. We followed two evaluation methods to assess the performance of these models: with fine-tuning the models on the proposed dataset by making the bottom layers trainable, and without fine-tuning. \textbf{Table-3} shows the difference in accuracy for each pretrained model using the two methods.

The results demonstrated the dataset's potential to achieve high-accuracy predictive models. For example, the DenseNet169 model attained a top-1 accuracy score of 88\%, and the ResNet-50 model achieved an accuracy score of 86\% across the different styles and categories, highlighting the dataset's robustness in training models that excel in classification tasks with nuanced visual differences.

\begin{table}[h]
\centering
\begin{tabular}{@{}lcc@{}}
\toprule
\textbf{Models} & \textbf{Inception Score $\uparrow$} & \textbf{FID} \\
\midrule
RAT-GAN \cite{ye2022recurrent}         & 4.04  & 24.02 \\
DC-GAN\cite{chen2016dcan}           & 3.38  & 30 \\
Projected-GAN\cite{sauer2021projected}    & 4.33  & 22.02 \\
\bottomrule
\end{tabular}
\caption{GAN MODELS PERFOMANCE}
\label{tab:classification}
\end{table}

\subsection{\textbf{Text to image}}
\begin{figure}[h]
    \centering
    \includegraphics[width=.9\linewidth]{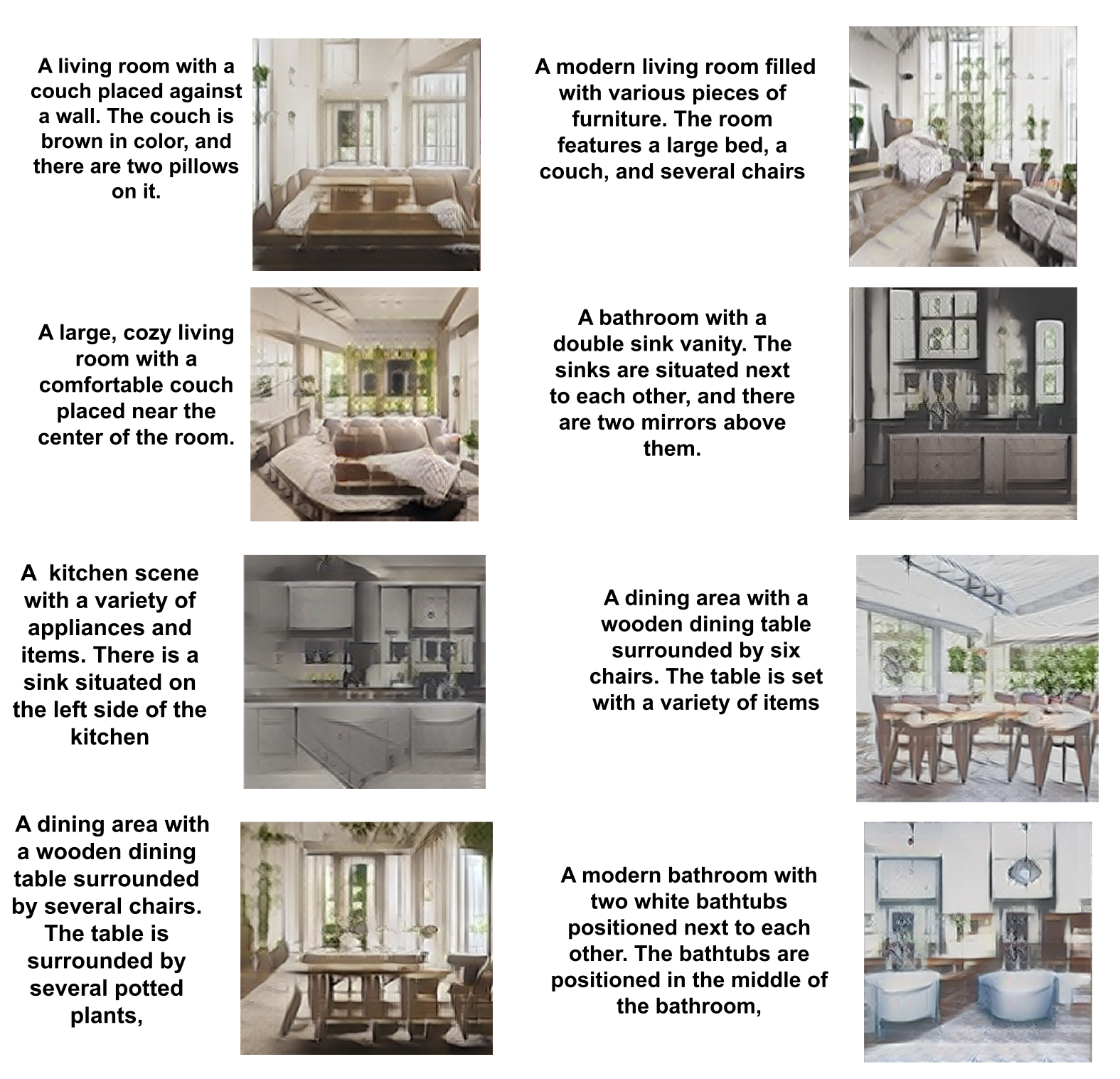}
    \caption{Text to Image Using RAT-GAN}
    \label{fig:enter-label}
\end{figure}

\subsection{Generation Benchmark}
For the image generation benchmark, we employed three distinct GAN models to thoroughly evaluate the capabilities of our interior design multimodal dataset: RATGAN, ProjectedGAN \cite{sauer2021projected}, and DCGAN \cite{chen2016dcan}. RATGAN \cite{ye2022recurrent} was specifically used for text-to-image generation, where the model synthesized interior design images from textual descriptions. This enabled it to translate detailed prompts into realistic images that accurately captured the stylistic nuances of the design styles. ProjectedGAN \cite{sauer2021projected}, on the other hand, was employed for image-to-image translation, transforming existing images within the dataset to reflect different styles or room categories. This model demonstrated an impressive ability to adapt and translate source images to new design contexts, creatively reimagining the interior of a given room in another style or category while maintaining stylistic consistency. DCGAN \cite{chen2016dcan}, another model used for image-to-image translation, synthesized high-quality images that adhered to the original dataset’s structural characteristics while offering inventive variations across different categories. The visual samples generated by all three models revealed the nuanced diversity of styles and categories within our dataset, confirming that our multimodal dataset is a rich resource for training GAN models capable of producing accurate, diverse, and creative representations of interior design. \textbf{Table-4} shows the inception score and the Frechet Inception Distance for the three models.

\begin{figure}[h]
    \centering
    \includegraphics[width=1\linewidth]{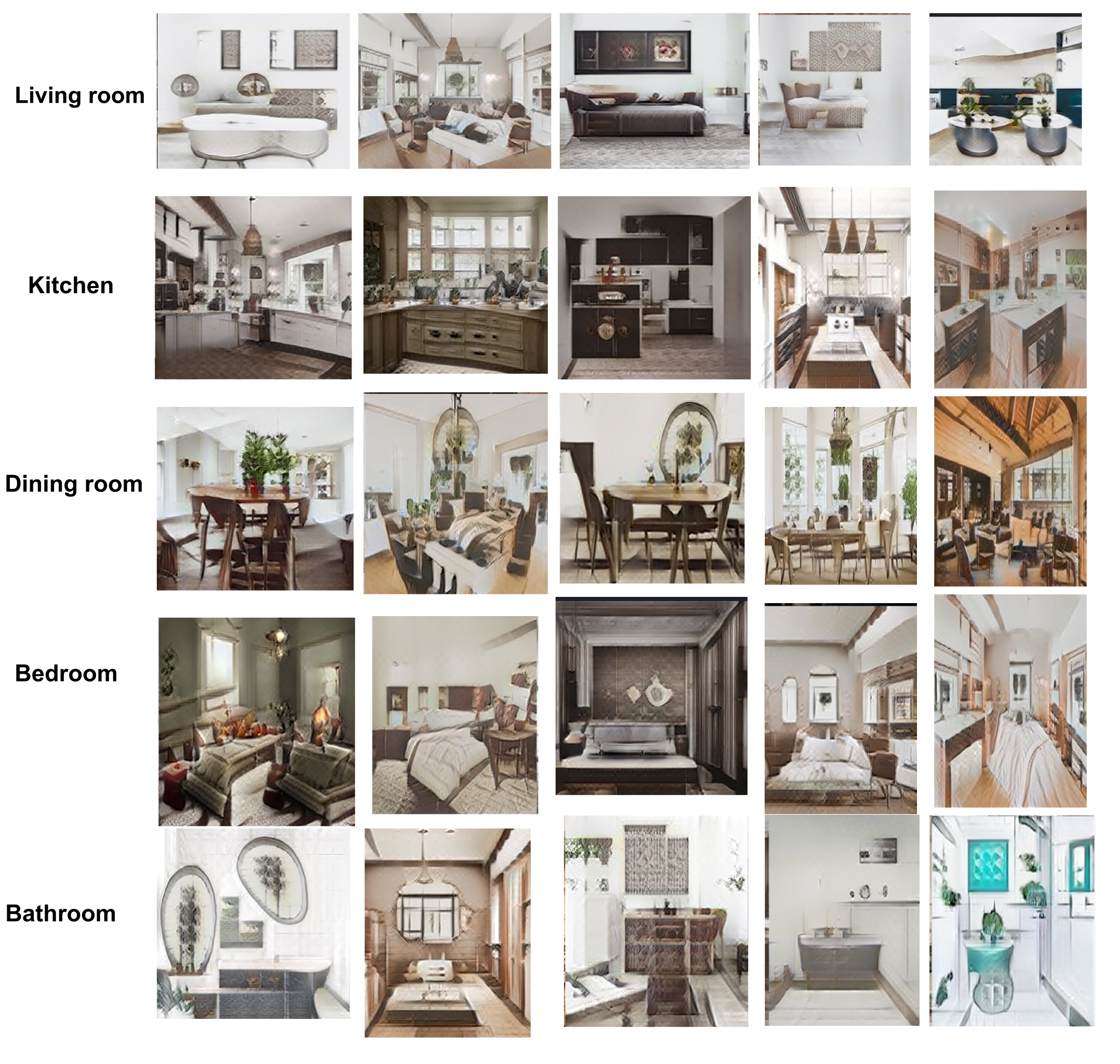}
    \caption{Image to Image using projected GAN}
    \label{fig:enter-label}
\end{figure}

\section{Conclusion }
In this paper, we introduced a novel interior design multimodal dataset that advances research and development in multimodal-related machine learning applications. Our benchmark experiments demonstrated the dataset's robustness and applicability, achieving high predictive accuracy for classification. Additionally, our exploration of GAN \cite{goodfellow2014generative} models—RATGAN, ProjectedGAN, and DCGAN—showcased their ability to generate accurate, diverse, and creative outputs, reflecting the original dataset's visual fidelity and variety. This comprehensive resource facilitates training and evaluating machine learning models in various tasks, including classification, image generation, and retrieval. this dataset will substantially accelerate research and applications in interior design, contributing valuable advancements to the fields of computer vision and natural language processing.

\section{Acknowledgment}
We extend our heartfelt gratitude to \textit{AiTech for Artificial Intelligence \& Software Development} (\url{https://aitech.net.au}) for providing the computational resources essential for our experiments. Their support has been crucial to the successful completion of this research.
.

\bibliographystyle{plain}
\bibliography{main}

\end{document}